\documentclass[conference]{IEEEtran}
\IEEEoverridecommandlockouts
\usepackage{cite}
\usepackage{amsmath,amssymb,amsfonts}
\usepackage{algorithmic}
\usepackage{graphicx}
\usepackage{textcomp}
\usepackage{xcolor}
\def\BibTeX{{\rm B\kern-.05em{\sc i\kern-.025em b}\kern-.08em
    T\kern-.1667em\lower.7ex\hbox{E}\kern-.125emX}}
\usepackage[]{graphicx} % omit 'demo' for real document
\usepackage{array}
\usepackage{mathtools}
\usepackage{amsmath}
\usepackage{setspace}

\def\mund#1{{\smallskip\noindent{\bf #1.}}}    
    
\begin{document}

\title{Patient ADE Risk Prediction through Hierarchical Time-Aware Neural Network Using Claim Codes}

\author{\IEEEauthorblockN{Jinhe Shi\IEEEauthorrefmark{1}, Xiangyu Gao\IEEEauthorrefmark{1}, Chenyu Ha\IEEEauthorrefmark{2}, Yage Wang\IEEEauthorrefmark{2}, Guodong Gao\IEEEauthorrefmark{3} and Yi Chen\IEEEauthorrefmark{1}}
\IEEEauthorblockA{\IEEEauthorrefmark{1}New Jersey Institute of Technology, Newark, NJ, USA\\
Email: {\{js675, xg77, yi.chen\}@njit.edu}}
\IEEEauthorblockA{\IEEEauthorrefmark{2}Inovalon, Bowie, MD, USA\\
Email: {\{cha, ywang2\}@inovalon.com}}
\IEEEauthorblockA{\IEEEauthorrefmark{3}University of Maryland, College Park, MD, USA\\
Email: ggao@rhsmith.umd.edu}}

\maketitle

\begin{abstract}
Adverse drug events (ADEs) are a serious health problem that can be life-threatening. While a lot of studies have been performed on detect  correlation between a drug and an AE, limited studies have been conducted on  personalized ADE risk prediction. Among treatment alternatives, avoiding the drug that has high likelihood of causing severe AE can help physicians to provide safer treatment to patients. Existing work on personalized ADE risk prediction uses the information obtained in the current medical visit. However, on the other hand, medical history reveals each patient’s unique characteristics and comprehensive medical information. The goal of this study is to assess personalized ADE risks that a target drug may induce on a target patient, based on patient medical history recorded in claims codes, which provide information about diagnosis, drugs taken, related medical supplies besides billing information. We developed a HTNNR model (Hierarchical Time-aware Neural Network for ADE Risk) that capture characteristics of claim codes and their relationship. The empirical evaluation show that the proposed HTNNR model substantially outperforms the comparison methods,  especially  for rare drugs.
\end{abstract}


\section{Introduction}\label{sec:Intro}

%% Importance of ADR discovery
\emph{Adverse Drug Events (ADEs)}, defined as ``an appreciably harmful or unpleasant event resulting from the use or misuse of a drug'' ~\cite{nebeker2004clarifying}, are  a serious health problem that can be life-threatening. According to FDA, the number of ADEs reported to FAERS (FDA Adverse Event Reporting System)  resulting in death and serious outcomes increase consistently \cite{sonawane2018serious,health2018}. Statistics \cite{health2020} show each year ADEs account for over 3.5 million physician office visits, an estimated 1 million emergency department visits, and approximately 125,000 hospital admissions. For inpatient setting,  ADEs account for an estimated 1 in 3 of all  hospital \emph{adverse events (AE)}  and affect about 2 million hospital stays each year.

%2. pre-marketing review and FAERS
 While pre-marketing review is conducted before any drugs are approved for marketing, it is insufficient for identifying all the potential ADEs due to  the limited sample size and duration of clinical trials. Post-marketing surveillance is critical for identifying ADRs. Although patient can report ADE through  voluntary and spontaneous report systems, such as FDA FAERS, the median under-reporting rate across 37 studies using a wide variety of post-marketing surveillance methods from 12 countries is 94\% according to an earlier study \cite{hazell2006under}.

 % EHR,  Claims => merits
There are increasing interests of using large-scale longitudinal clinical data, EHRs,  associated  clinical notes, as well as claims  data,  for studying ADEs. Such data  contain rich and accurate information about patients’ health status, their treatment plan and clinical outcomes. Since such data is generated as part of medical practices, without relying on patient self-reporting, it is available in large-scale with high quality.

 % category 1: drug-ADR detection
The studies can be categorized into two types:  ADE detection and  personalized ADE risk prediction. The goal of ADE detection is to identify the correlation or causal relationship between a target drug and an observed AE.  Some  use statistic methods  such as the disproportionality analysis \cite{montastruc2011benefits,evans2001use,lependu2013pharmacovigilance}, others use machine learning methods such as support vector machines, random forests and neural networks \cite{henriksson2015identifying,wang2015method}.
%Their ground truth is either labeled by humans or based on a database of known ADEs, such as SIDER.
%
Besides detecting drug-AE correlation  on the whole patient population, there are also studies  on ADE risk stratification which assesses the  correlation on patient populations defined by their demographics \cite{haerian2012detection}.

%% category 2: personalized adr risk prediction
In contrast of ADE detection for  population, the studies on personalized ADE risk prediction assess  the likelihood of individuals to experience an AE  based on individual characteristics and clinical history. Indeed different patient may have different AE outcomes even taking the same drug. Among alternative drugs for treatment, avoiding the one that has high likelihood of causing severe AE can help physicians to provide safer treatment to patients, as a form of personalized treatment. There are only a few works addressing the problem \cite{mcmaster2019machine,bos2018prediction}. They take as input patient demographic information and clinical information of the current hospital visit.

%% importance of considering medical history, our uniqueness
What is lacking in the literature is  to consider patient medical history in addition to  the current visit information to make personalized prediction for ADE risks.
Medical history better reveals each patient's unique characteristics, as well as the drugs and treatments taken in the past, which may interact with the current treatment to induce AE \cite{liu2017data, jiang2015mining}.

%% challenge of considering medical history
However, patient medical history data is often not readily available and is difficult to process. First, patient may be seen at multiple healthcare centers that do not share patient data in their EHR system. 
Second, patient self-reporting medical history may not be accurate or comprehensive. 
Furthermore, processing large-scale  longitudinal medical history data, which contains diverse type of clinic information,  poses technical complexity

% use claim data: application / motivation
% unique advantage of claims data
The goal of this study is to assess personalized ADE risks that  a target drug may induce on a target patient, based on patient medical history recorded in claims, which we acquired access via collaboration with Inovalon, a healthcare analytics company.
Our findings can be used by Medicare/Medicaid and health insurance company to provide assistance to healthcare professionals to identify safe treatment plan. 

Claims  data provide valuable information about patients. It contains  the information  about  diagnoses, drugs taken, related medical supplies, treatment procedures,  besides billing information, for each patient encounter. While a patient may receive healthcare from multiple providers, and have their medical information scattered in multiple EHR systems, claims data  effectively  records  a patient’s interactions across different healthcare systems and thus  provides longitudinal and accurate data in the continuum of a patient's health care history \cite{stein2014use}. 

%(64,070 unique claim codes)

However, there are several technical challenges that must be addressed. 
The first challenge is how to capture the ``meanings'' of claim codes. There are over 64K unique claims code in the data,  belong to nine different types.
We make an analogy between claim codes and words, and between claim history and documents. Then we propose to use word embedding methods in Natural Language Processing (NLP) to generate embedding for claim codes, so that claim codes that are used in similar ways are represented with similar vectors, naturally capturing their meanings. 

The second challenge is how to model  patient medical claim history. A patient's claim history consists of encounters and each encounter consists of claim codes. The relationship of claim codes within an encounter is different from that of claim codes in different encounters. This present a unique challenge, as exiting work does not consider patient's medical history but only the current medical visit.
To model patient's claim history, we  propose a HTNNR model stands for Hierarchical Time-aware Neural Network with drug-code Representation. The first layer neural network encodes claim codes within an encounter into vectors, and the second layer neural network represents the claim history with a sequence of encounters into vectors.
Then we  propose to use a bi-directional neural network model to capture the un-ordered relationship among claim codes within an encounter. We further propose to use time-aware deep learning model to capture not only the sequential but also the temporal relationship among encounters.

%% main contributions. 
The  contributions of our work include the following. First, to the best of our knowledge, this the first study that uses patient claim history to make  personalized prediction on  drug-induced  ADE risks. 
Second, we have made several technical contributions. We proposed claims code embedding, a hierarchical neural network model to capture patient claim history, and  drug-claim code representations. We also used different neural network models for encounter representation and for claim history representations.  
Finally, extensive evaluation on about 500k patients demonstrates  effective prediction performance and high efficiency of our proposed approach.

The rest of the paper is organized as follows. Section \ref{sec:rel} discusses the related work. Section \ref{sec:prob} presents the problem statement and data overview. Section \ref{sec:method1} and Section \ref{sec:method2} presents the two methods for patient ADE risk prediction. Experimental results are presented in Section \ref{sec:exp}. Section \ref{sec:futurework} concludes the paper.

\begin{table*}[ht]
\caption{Description of different claim codes}
\centering  
\begin{tabular}{|p{3cm}|p{12cm}|}\hline
Code Type        &Description \\\hline
ICD    & International Statistical Classification of Diseases (ICD) codes capturing  diseases,  symptoms, abnormal findings, complaints, etc.  It includes diagnosis codes (ICD10DX and ICD9DX) and procedure codes (ICD9PX and ICD10PX). \\    \hline
CPT   & report medical, surgical, and diagnostic procedures and services to entities such as physicians, health insurance companies and accreditation organizations      \\ \hline
POS     &Place of Service (POS) Codes are two-digit codes placed on health care professional claims to indicate the setting in which a service was provided.      \\\hline
GPI &The Generic Product Identifier (GPI) is a 14-character hierarchical classification system that identifies drugs from their primary therapeutic use down to the unique interchangeable product regardless of manufacturer or package size.   \\\hline
TOB     &Type of bill codes (TOB) identifies the type of bill being submitted to a payer. TOB codes are four-digit alphanumeric codes that specify different pieces of information on claim form      \\\hline
REVENUE &Revenue Codes are descriptions and dollar amounts charged for hospital services provided to a patient.        \\\hline

HCPCS &The Healthcare Common Procedure Coding System (HCPCS) is a collection of codes that represent procedures, supplies, products and services which may be provided to Medicare beneficiaries and to individuals enrolled in private health insurance programs.       \\\hline
DISCHARGE & Identify where the patient is at the conclusion of a health care facility encounter (a visit or an inpatient stay) \\\hline

LOINC &Logical Observation Identifiers Names and Codes (LOINC) is a database and universal standard for identifying medical laboratory observations \\\hline

\end{tabular} \label{tab:code}
\end{table*}

\section{Related Work}\label{sec:rel}

Studies on  ADEs can be categorized into ADE detection on population,  personalized AE risk prediction, and  prediction of ADE outcome intensity (e.g. hospitalization and mortality).

The goal of ADE detection is to identify the correlation or causal relationship between a target drug and an observed AE, using  statistical methods or machine learning methods.
Some studies  applied association rule mining methods for ADE detection \cite{wang2012exploration,reps2014refining}. 
Disproportionality analysis are widely used for ADE detection from various data sources, such as  EHR data \cite{lo2013mining,harpaz2013combing},  clinic notes \cite{lependu2013pharmacovigilance}, and clinical trials \cite{dias2015role}.

Disproportionality analysis  is based on the contrast between observed and expected numbers of co-occurrences, for any given combination of drug and AE, to detect possible causal relations between drugs and AEs. 
It, however, does not consider context features, which are rich in unstructured clinical notes.  Various Natural Language Processing (NLP) and machine learning techniques have been applied on clinic notes to detect drug-AE association, using expert-labeled ground truth.
\cite{wang2015method,henriksson2015identifying} extract multiple features like drug and AE frequency and co-mention frequency from clinical notes and use machine learning methods like support vector machine  and random forest to detect drug-AE correlation. 
 \cite{chapman2019detecting, dandala2019adverse} start with a named entity recognition  module based on Conditional Random Fields  to extract medical entities relevant to ADEs from clinical notes, and then use random forest and neural networks, respectively, as the relation classification model.
Little has been studied on using claims data for ADE detection. 
 \cite{kwak2020drug} use ICD codes and GPI drug code in claims data (see Table  \ref{tab:code} for description of the code) as input and design a graph neural network model to construct a drug-disease graph for ADE detection. They first embedded disease codes and drug codes into a graph, respectively, then the merged drug and disease graph is fed into a graph neural network for ADE detection. They used the SIDER database as the ground truth for ADEs.
Besides detecting drug-AE correlation  on the whole patient population, there are also studies  on ADE risk stratification which assesses the drug-AE correlation on patient populations defined by their demographics \cite{haerian2012detection}.

There are only a few studies in the category of personalized AE risk prediction. 
Since  AE risks of different patients are different, even for the same drug, these studies make risk predictions based on the individual patient's characteristics from clinical data.
 \cite{bos2018prediction} develops a logistic regression model to predict the risks of AEs of in-patients based on the patient features and the medical conditions during this hospital stay. They used multiple patient characteristics like gender and age as features, also extracted some features from current medical conditions like the number of medications and the list of drugs taken.
  \cite{mcmaster2019machine} takes  clinical features as input, such as  ADE indication codes, primary diagnosis code and length of the hospital stay to predict in-patient ADE risks. 
% by classifying them into four ADE risk probability class like possible, definite. 
They used multiple machine learning models like random forest and support vector machines. 
%Both of them used human labeled ground truth. 
Both make ADE risk prediction based on the information of the current hospital stay.
Being most related to this category of studies, our work takes as input a patient's longitudinal medical history, not just the current medical encounter. Also, we consider AE risks induced by target drugs (perhaps due to interaction with other drugs or medical conditions), whereas existing studies consider AE in general. The dataset used in our studies is claims data.

Unlike studies on personalized AE risk prediction, which predict the likelihood of a specific AE to occur, there are also studies on predicting the likelihood of hospitalization and mortality of a patient, due to outcomes of unspecified AEs. Both of them are using the patient medical data from FAERS.
 \cite{islam2018detecting} proposed a hybrid model  to  predict the  outcomes of ADEs, based on patients’ demographic data, such as age and gender, and drug-taken information, such as the route of the drug intake  and  whether the adverse reaction subsided when drug in-take was terminated.  
 \cite{valeanu2020development} developed a system that takes patient demographics, drugs,  relevant diseases in pathology  as input, and  outputs ADE risk outcome assessment.

 \section{Problem Statement}\label{sec:prob}
In this section we present the data description and the problem definition. 

% \yc{you used ADR and ADE in the paper. use one term only. I prefer ADE. but if the claims code is for ADR, then use ADR consistently.
%  }
%  \jh{I checked related work, the mainly use ADE when mentioned claim codes. I will use ADE in this paper. But I also found some work using ADR. I'm not sure. We had ADR discovery in online health forums, do you think it is better to consistent with that work?}
%  \yc{use ADE. for your dissertation, you can change all to ADE}

\subsection{Data Description} 
\label{sec:data}
The input data is  medical claim history for a set of patients. Each claim history is composed of  a sequence of encounters, and each encounter has a sequence of claim codes, as illustrated in  Figure \ref{fig:probStatement}. At an encounter, a medical treatment and/or evaluation and management services are provided.
There are nine different types of claim codes,  which provide information of medical diagnoses (ICD), procedures and services (CPT, LOINC), setting where services are provided (POS), drug information (GPI), billing (TOB, REVENUE, DISCHARGE), and codes for Medicare and private health insurance program users (HCPCS). Table \ref{tab:code} shows a description about these code types.

\begin{figure}[ht]
\centering
\includegraphics[width=0.7\linewidth]{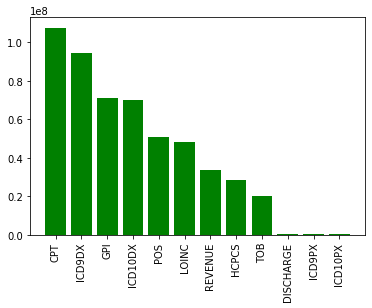}
\caption{Distribution of Claim Code Occurrences}
\label{fig:code_category}
\end{figure}

\begin{figure}[ht]
\centering
\includegraphics[width=0.7\linewidth]{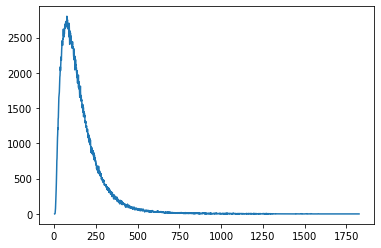}
\caption{Distribution of Number of Encounters Per Patient}
\label{fig:encounter_length}
\end{figure}

The data used in empirical evaluation was provided by Inolvaon, a technology company providing cloud-based platforms empowering data-driven healthcare. It contains the claims data of 500k patients for a duration of 2015-2019. There are 64,070 unique claim codes.
Figure \ref{fig:encounter_length} shows the distribution of the number of encounters a patient has. We can see that most patient has less than 500 encounters, and the average number of encounters per patient is $158.7$. 
The average number of claim codes per patient is $1052$ and the average number of claim code per encounter is $6.6$.
Figure \ref{fig:code_category} shows the number of claim code occurrences of each category.

% \begin{figure}[ht]
%   \caption{Data Statistics Analysis}
%   \label{fig:data}
% \begin{subfigure}[t]{0.48\linewidth}
% \imagebox{25mm}{\includegraphics[scale=0.3]{fig/codeDis.png}}
% %   \includegraphics[width=\linewidth]{fig/codeDis.png}
% %   \caption{Claim Code Category Occurrence Distribution}
%   \label{fig:code_category}
% \end{subfigure}%
% \vspace{5.0mm}
% \begin{subfigure}[t]{0.48\linewidth}
% \imagebox{25mm}{\includegraphics[scale=0.3]{fig/patientEncounterLength.png}}
% %   \includegraphics[width=\linewidth]{fig/patientEncounterLength.png}
% %   \caption{Distribution of Number of Encounters Per Patient}
%   \label{fig:encounter_length}
% \end{subfigure} 
% \end{figure}

\subsection{Problem Definition}
Now we formally define the problem.

We model the problem as a classification task. A patient's claim history is composed of a sequence of encounters, denoted as  $P=\{e_1, e_2, \dots\}$. Each encounter $e_i$ is composed of a sequence of claim codes, $e_i=\{x_1, x_2, \dots\}$.  Consider a  list of target ADEs, and  a target drug $d$. $y \in \{-1,1\}$ is the classification label, where $y=1$ indicates that drug $d$ induced at least one ADE in the target ADE list on this  patient, and otherwise $y=-1$. For a set of patients who took drug $d$, their claim histories before taking $d$ along with their corresponding labels are used to train the classification model.
For a target patient who has not taken  drug $d$, the  model takes his claim history so far to predict the label, i.e. whether he will experience an ADE in the target ADE list if  taking $d$ now.

\begin{figure*}[ht]
  \centering
  \includegraphics[width=\linewidth]{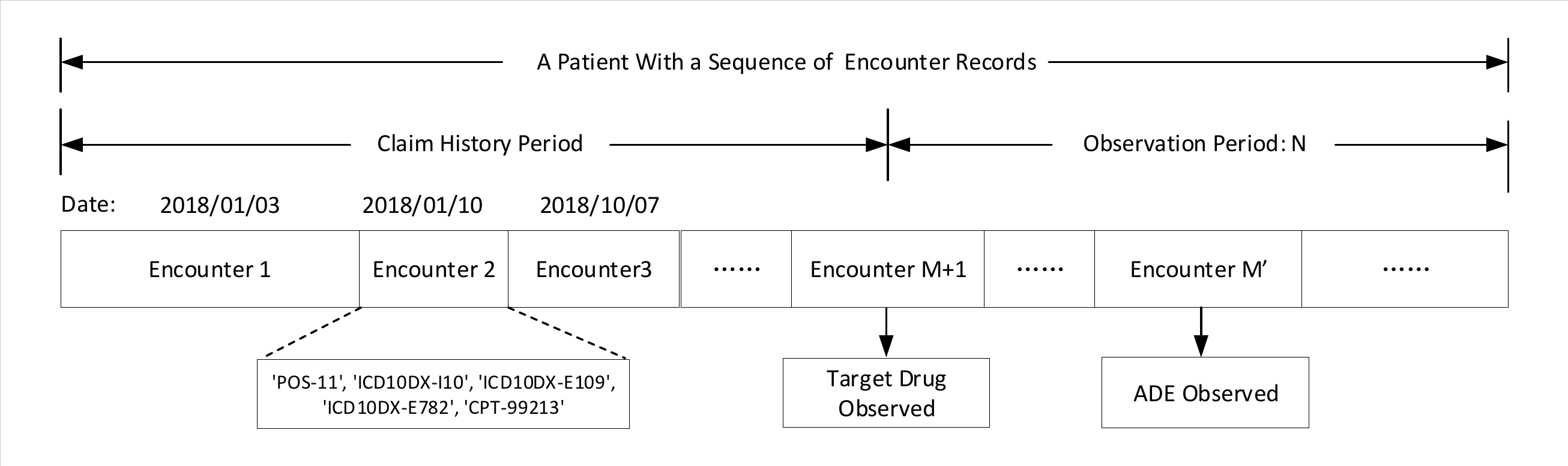}
  \caption{Claim History Illustration}
  \label{fig:probStatement}
\end{figure*}

\mund{Identifying ADEs from Claim History}
First, we identify ADEs from claims code. Based on literature, an ADE can be identified from  the claim codes  by concurrent  presence of selected diagnosis codes and   selected indication codes  \cite{walter2017impact}.  
Diagnosis codes are part of the ICD codes as shown in Table \ref{tab:code}.
An indication code  is a special type of diagnosis code indicates that a patient experienced an ADE  \cite{walter2017impact,hohl2014icd}.
 Following existing work \cite{hohl2014icd}, we use four categories of indication codes as shown in  Table \ref{tab:indicationCodes} and their corresponding ICD codes.  For example, ICD code ``T46.9" represents ``Other and unspecified agents primarily affecting the cardiovascular system" is an indication code, indicating an ADE related to cardiovascular system. 
 %Finally, we get total 532 indication codes
If a diagnosis code and an indication code co-occur in an encounter,  we consider an ADE occurs and the diagnosis code gives the information of the AE. For example: if a diagnosis code ``I42.7" (Cardiomyopathy due to drugs and other external agents) and ``T46.9'' both occurs in an encounter, then ``I42.7" represents an ADE.

The diagnosis codes (ICD codes) of the target ADEs and the GPI  code of the target drugs are  input of the problem. 
We consider that  a target drug induces a target ADE experienced by a patient if the ICD code corresponding to 
the target ADE and one of the indication codes in Table \ref{tab:indicationCodes}  are found in the same encounter within  time period $N$  after taking the  drug, but not found in the claim history before taking the drug, 
Specifically, as illustrated in Figure \ref{fig:probStatement}, suppose a patient starts to take a target drug from encounter $e_{M+1}$. If there is no target  ADE found before encounter $e_{M+1}$ but is recorded in encounter $e_{M^{'}}$ along with an indication code, and the time duration between $e_{M+1}$ and $e_{M^{'}}$ is less than $N$, then we consider the target drug induces  this ADE. We can also use other approaches to generate ground truth, such as human labeling.

Note that it is possible that an ADE is a result of drug-drug interaction \cite{alvim2015adverse}. In other words, some time multiple drugs together induce to an ADE.  For any of these  drug is a target drug, for this drug  the corresponding claim sequence is labeled positively.

Also, $N$ is  considered the effective time of a drug to cause AE. Currently $N$ is set to be 3 months for all the drugs. Different values of $N$ can be used for different drugs based on the drug characteristics when the information becomes available.

% \begin{table*}[h]
% \caption{Indications Codes for Adverse Drug Events}
% \centering  
% \begin{tabular}{|p{3cm}|p{8cm}|p{3cm}|}\hline
% Indication Category        &Description  &Example Codes\\\hline
% A1   & The ICD-10 code description includes the phrase 'induced by medication/drug' & J70.2\\    \hline
% A2   & The ICD-10 code description includes the phrase 'induced by medication
% or other causes'     &T88.7\\ \hline
% B1       &The ICD-10 code description includes the phrase 'poisoning by medication'.   & T36 - T50   \\\hline
% B2      &The ICD-10 code description includes the phrase 'poisoning by or harmful use of medication or other causes'    &X44\\\hline
% \end{tabular} \label{tab:indicationCodes}
% \end{table*}

\begin{table}[h]
\caption{Indication Codes for Adverse Drug Events}
\centering  
\begin{tabular}{|p{0.5cm}|p{5cm}|}
\hline
\multicolumn{1}{|l|}{\textbf{Indication Category }} &
\multicolumn{1}{l|}{\textbf{Description}}
\\ \hline
A1   & The ICD-10 code description includes the phrase 'induced by medication/drug' \\    \hline
A2   & The ICD-10 code description includes the phrase 'induced by medication
or other causes'     \\ \hline
B1       &The ICD-10 code description includes the phrase 'poisoning by medication'.     \\\hline
B2      &The ICD-10 code description includes the phrase 'poisoning by or harmful use of medication or other causes'   \\\hline
\end{tabular} \label{tab:indicationCodes}
\end{table}

\section{First Attempt}\label{sec:method1}
\label{sec:first}

\begin{figure}[ht]
  \centering
  \includegraphics[width=\linewidth]{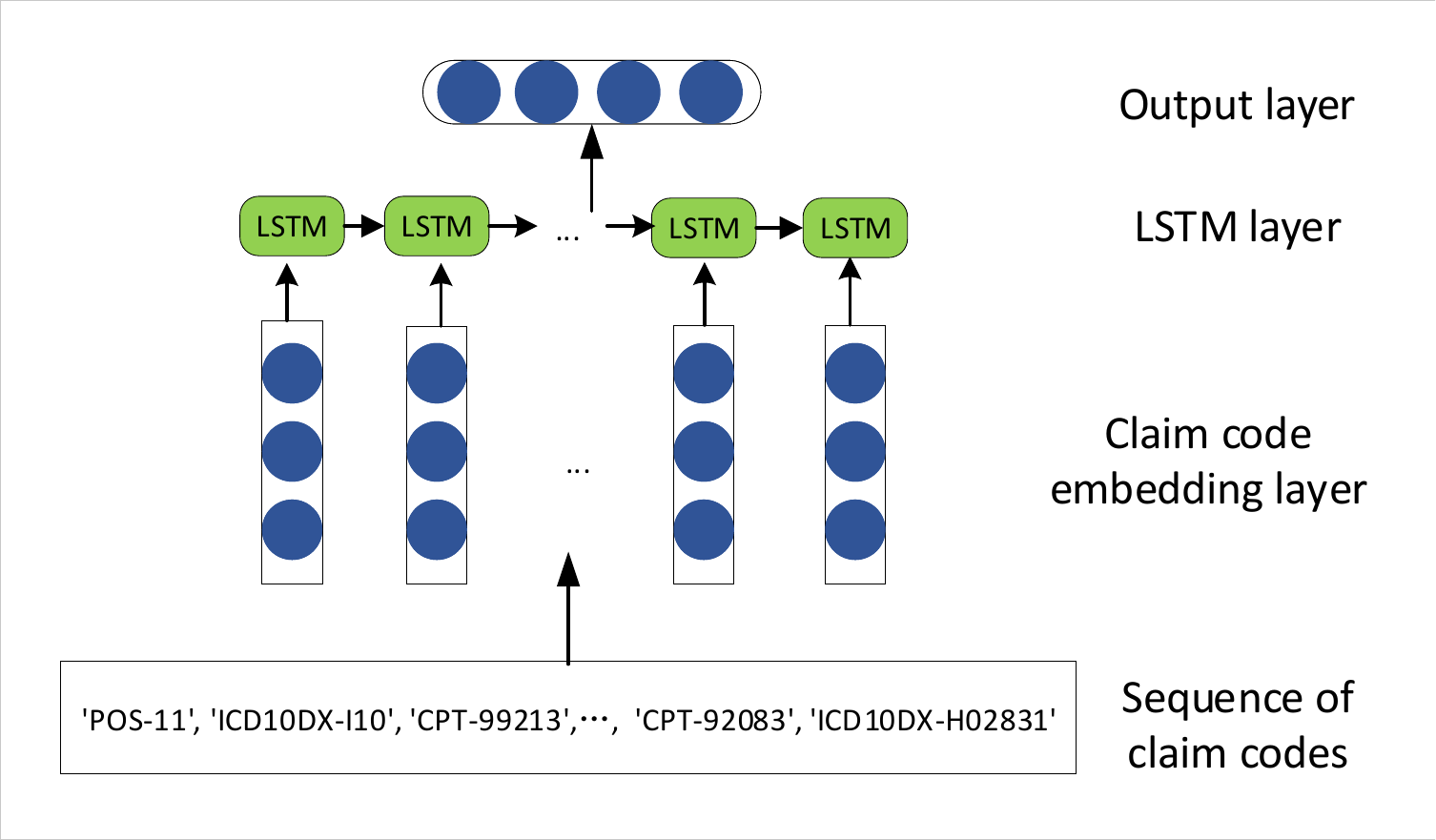}
  \caption{The architecture of First Attempt Method}
  \label{fig:method1}
\end{figure}

Since patient's medical claim history consists of  a sequence of claim codes which
encode  medical diagnoses, procedures and services conducted, drugs taken and so on. 
Intuitively, the problem can be modeled as a sequence classification problem.
Figure \ref{fig:method1} shows a system architecture. The input is the patient's medical claim history represented as a sequence of claim codes.cThen each claim code is represented as an embedding vector. A deep learning model, Long Short-Term Memory (LSTM), is then used to learn the dependency between the claim codes in order to make the prediction whether this patient will experience a target ADE if taking a target drug.

\mund{Claim Code Embedding}
The claim code embedding layer  generates a vector for each claim code that captures the characteristics of codes and the relationship among codes. Word embedding is widely used in deep learning based NLP techniques. Using dense and low-dimensional vectors to encode words bring computational benefits to downstream neural network model processing. Learned based on word usage, word embedding represents  words that are used in similar ways using similar vectors, naturally capturing their meaning.
We make the analogy that each claim code corresponds to a word, an encounter corresponds to a sentence,  and a patient claims history corresponds to a document.  The usage of claim code  indicate their correlation, just like the usage of words in text.
We use  a popular word embedding method in NLP, the skip-gram model \cite{guthrie2006closer}. It takes as input  the collection of all patient's claim code sequences and generates
 a low dimensional, continuous and real-value vector for each claim code as its embedding.

\mund{Sequence Classification with LSTM} After using an embedding to represent  each claim code, the sequence of   claim code embedding is fed into a deep learning model to learn the claim code dependencies. We  identify all patients in the training data who took a target drug.
 These patients' sequences of claim code embedding before  taking the target drug, and the corresponding labels of whether a target ADE is observed within the L time period after taking the drug in the claim history are used to train the model. 
 The trained model then  predicts the label  of each patient in the test data, based on his/her claim code embedding sequence so far.

In contrast to Convolutional Neural Network (CNN), Recurrent Neural Networks (RNN) are designed for sequence prediction problems \footnote{The performance evaluation of CNN and and several feature-based machine learning methods are presented in Section \ref{sec:exp}.}. However, it suffers the problem of gradient vanishing or exploding \cite{hochreiter1997long}, where gradients may grow or decay exponentially over long sequences. This makes it difficult to model long-distance correlations in claim code sequences. Recall that  the average number of claim codes per patient is 1052.

We proposed to use LSTM networks instead, which are designed to  overcome  the vanishing gradient problem and to efficiently learn long term dependencies. LSTMs accomplish this by keeping an internal state that represents the memory cell of the LSTM neuron. This internal state controls the information flow through the cell state. 

The new cell state $c_j$ and the output $h_j$ can be calculated as:
\begin{equation}
\label{eqn:01}
c_j= f_j\odot c_{j-1} + I_j\odot tanh(W_{c}[F_j, h_{j-1}]+ b_c)
\end{equation}
\begin{equation}
\label{eqn:02}
h_j= o_j\odot tanh(c_j)
\end{equation}
where $I_j$, $f_j$ and $o_j$ denote input, forget and output gate, respectively.
Finally  the output layer uses a softmax function on the  vector generated from the LSTM layer to make a prediction.
This approach is referred as \emph{LSTM} in the rest of paper.

\section{HTNNR Model}\label{sec:method2}

After presenting the LSTM method in Section  \ref{sec:first}, now we discuss several characteristics of patient claim code history and propose a novel model named as \emph{HTNNR Model} stands for Hierarchical Time-Aware Neural Network for ADE Risk. 

\subsection{A Hierarchical Neural Network}
 The LSTM method models patient claim history as a sequence of claims code. However, this approach may not accurately capture the relationship between the claims codes.  Recall that the claim  history  actually consists of a sequence of encounters, each of which contains a sequence claim codes. There are two observations.
First, the number of claim codes in different encounters can have big variation. 
For instance, consider three encounters illustrated in Figure \ref{fig:probStatement}. The first encounter represents a hospital stay, with  30 claim codes. The next encounter  represents a follow-up with a specialist, with only four  claim codes. The third encounter represents a visit to a primary care doctor for a flu with another four claim codes.  The LSTM model ignores the encounter information, but just considers the claim code sequence where code relationships are reflected by their distances.
In this example, the 1st code  and the 30-th code are considered less related since their  distance is 29, despite that they actually belong to the same encounter. On the other hand,  the 30-th code and the 35-th one are considered as closely related since their distance is only 5. However, they actually are  two encounters apart, and are not semantically closely related. 
The second observation is that  the claims code within an encounter are actually not ordered, collectively describing an encounter event.

\begin{figure*}[ht]
  \centering
  \includegraphics[width=0.9\linewidth]{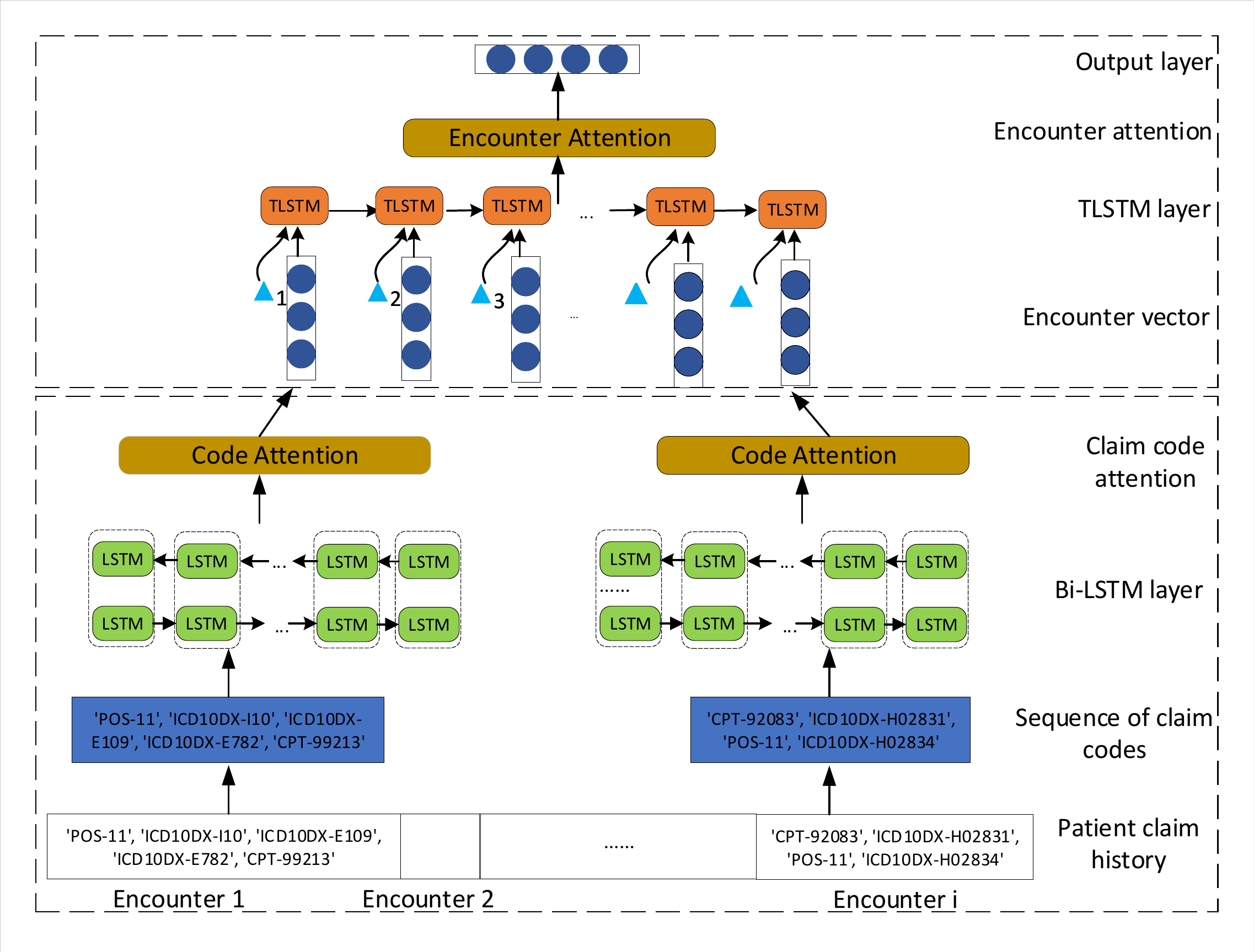}
  \caption{Architecture Overview of HTNNR}
  \label{fig:HAttention}
\end{figure*}

Based on this observation,  we propose a hierarchical framework to model the input data, as shown in Figure \ref{fig:HAttention}. The first layer in framework generates a vector for  each  encounter, called \emph{Encounter Representation}.  The second layer in the framework takes the sequence of encounter vectors as input and outputs an embedded vector for each patient's claim history, referred to as \emph{Claim History Representation}. This framework better captures the claim code relationships. Now we discuss these two layers in term.

\subsection{Encounter Representation}
The Encounter Representation takes the patient claim history as input. It has two components: a Bi-LSTM layer and a claim code attention layer. We discuss each in turn.

\mund{Bi-LSTM Representation for Encounters}
Recall that the LSTM method discussed in Section  \ref{sec:method1} consider claim history as a sequence of claim codes.
However claim codes in an encounter do not have sequential order, but are a set of codes that collectively record an encounter event. Based on this observation, we propose to use Bi-directional Long Short Term Memory (Bi-LSTM)  \cite{graves2013speech} to generate a representation of  claim codes in an encounter, which are unordered. Both previous codes and following codes within an encounter are considered by Bi-LSTM
to model  code dependencies. The output of the $j^{th}$ claim code in an encounter is calculated as:
\begin{equation}
\label{eqn:03}
h_j = \stackrel{\rightarrow}{h_j}\oplus \stackrel{\leftarrow}{h_j},
\end{equation}
where $\oplus$ is an concatenation operation.

\mund{Claim Code Attention}
Not all claim codes contribute equally to the semantic representation of an encounter. Attention neural networks have recently demonstrated success in document classification by learning the weights of words \cite{yang2016hierarchical}. 
Hence, we apply the attention mechanism to set weights of claim codes, so that the model can focus on claim codes that are important to capture the semantics of an encounter. 
The encounter representation $\mathbf v_e$ is formed by a weighted sum of the vectors generated by Bi-LSTM.
\begin{equation}
\label{eqn:08}
E = tanh(\mathbf H)
\end{equation}
\begin{equation}
\label{eqn:09}
\alpha= softmax(w^\mathrm{T}E)
\end{equation}
\begin{equation}
\label{eqn:10}
\mathbf v_e = \mathbf H \alpha^\mathrm{T}
\end{equation}
Here  $\mathbf H$ is a matrix consisting of  vectors $\left[h_1, h_2, ...,h_T\right]$ that the Bi-LSTM layer produces, where $T$ is the input length.
 $w$ is a trained weight vector and $w^\mathrm{T}$ is a transpose. 

\subsection{Claim History Representation}
Given the encounter vectors $\mathbf v_{e_i}$
output by the Encounter Representation layer for every encounter $e_i$ in a claim histor, now we discuss how to generate vector for each patient's claim history.

One intuitive way is to use a LSTM model on the sequence of encounter vectors to generate a  claim history vector. Indeed the sequential order of encounter indicate the temporal order of the encounter events.  However, LSTM does not capture the time differences among the encounters.
 Referring to Figure  \ref{fig:probStatement}. The  first two encounters are 7 days apart, 
with  the second encounter being a follow-up visit of a surgery preformed in the first encounter. The time between the second and the third encounter is 9 months, with the third encounter being  a visit to a primary care doctor for a flu. As we can see from this example, two adjacent encounters that has a small time lap often  refer to  closely related medical issues. On the other hand,  two adjacent encounters that are a long time  apart  likely refer to  unrelated medical issues.  In this case, the previous encounter has less importance to the semantics of the current encounter. 
Thus, sequential order itself is inadequate to capture the relationship between encounters, we should also consider the actual time differences.

We propose to use a Time-aware LSTM (TLSTM)  \cite{baytas2017patient} to generate a claim history vector from  the sequence of encounter vectors for each patient. For each encounter, we consider not only its claims code, but also its timestamp. 
The major component of the TLSTM layer is the subspace decomposition applied on the memory of the previous time step. 
The short-term memory is adjusted proportionally to the amount of time span between two patient encounters.
\begin{equation}
\label{eqn:11}
g(\Delta_i)= 1/\Delta_i
\end{equation}
\begin{equation}
\label{eqn:12}
\hat c_{i-1}= c_{i-1} * g(\Delta_i) 
\end{equation}
\begin{equation}
\label{eqn:13}
c_{i-1}^* = c_{i-1}^L + \hat c_{i-1}
\end{equation}
Here $\Delta_i$ is the time span between encounter $e_i$ and encounter $e_{i-1}$,  $c_{i-1}$ is short memory in LSTM, $\hat c_{i-1}$ is the adjusted short memory by considering time span. $c_{i-1}^*$ is the final adjusted previous memory that combines the normal long term memory $c_{i-1}^L$ and the adjusted short term memory.
As we can see, if the gap between encounter $e_i$ and $e_{i-1}$ is large, which means there is no new information recorded for the patient for a long time,  the dependence on the short-term memory does not play a significant role in the prediction of the current output. 

In this way, the final cell state in Equation \ref{eqn:01} is changed to:
\begin{equation}
\label{eqn:14}
c_i= f_i\odot c_{i-1}^* + I_i\odot tanh(W_{c}[F_i, h_{i-1}]+ b_c)
\end{equation}

The patient claim history vectors are calculated from  encounter vectors  as the following:
\begin{equation}
\label{eqn:15}
h_i= TLSTM(\mathbf v_{e_i}, \Delta_i), i \in [1,M]
\end{equation}

Here $\mathbf v_{e_i}$ is the encounter representation for encounter $e_i$,
$M$ is the number of encounters before the target drug taken. $\Delta_i$ is the elapsed time between encounter $e_i$ and $e_{i-1}$.

Finally  the  patient claim history vector is fed into an attention layer to learn the  importance of different encounters to make the prediction whether the target patient will experience a target ADE.

\section{Experiment} \label{sec:exp}
We implemented the proposed method, referred as \emph{HTNNR}. 
We have conducted extensive experiments to empirically evaluate the HTNNR model using real-life claims data. We start with discussing the model implementation, evaluation setting  and comparison methods. Then we present the empirical evaluation results.
%the overall method performance. To analyze the method efficiency, we evaluate the performance of Drug-Claim Code Representation, Drug Selection and Patient Medical History Length

\subsection{System Implementation}
HTNNR is implemented using Python and the Hierarchical Attention model is implemented using Keras with Tensorflow backend. The experiments are run on a 20-core computer server. 
 Existing work indicates that a large batch size may alleviate the impact of noisy data, while a small size sometimes can accelerate of convergence \cite{hoffer2019augment}. We varied the batch size in experiments,
 and  set the training batch size to 256 considering the trade-off of performance and the consumption of training time and memory.
% e.g., 128 and 512, but no significant differences in model performance. 
To train ADE classification, we use binary cross-entropy as the loss function. The optimizer we adopted is Adaptive Moment Estimation (Adam) which can achieve fast gradient descent \cite{kingma2014adam}. We use validation-based early stopping to obtain the models that work the best with the validation data. The model with the minimum validation error are saved and used to make prediction  the testing data.

\subsection{Evaluation Setting}\label{sec:training}
The data we used is provided by Inovalon. Inovalon's $MORE^2$ Registry dataset contains 500K patients. Each patient contains a sequence of encounters and each encounter contains a sequence of claim codes, with statistics presented in Section \ref{sec:data}.

\mund{target Drugs} 
We evaluated our proposed methods on 10 randomly selected drugs among all drugs, each of which  has been taken by more than 20K  patients in the dataset. Table \ref{tab:DrugSelection1} shows the GPI, description and the number of patients taking the drug.   

\begin{table}[ht]
\small
\caption{Target Drugs}
\centering  
\begin{tabular}{|p{2.2cm}|p{3cm}|p{1.8cm}|}
\hline
\multicolumn{1}{|l|}{\textbf{Drug GPI code}} &
\multicolumn{1}{l|}{\textbf{Description}}&
\multicolumn{1}{l|}{\textbf{Patient Population}}
\\ \hline
GPI-5818002510      &Duloxetine HCl          &22616            \\\hline
GPI-3610003000      & Lisinopril             &124716            \\\hline
GPI-4927006000      & Omeprazole             &138152            \\\hline
GPI-3400000310      &Amlodipine Besylate     &127326           \\\hline
GPI-4220003230      &Fluticasone Propionate  &106106          \\\hline
GPI-3320003010      &Metoprolol Tartrate     &75561           \\\hline
GPI-3615004020      &Losartan Potassium      &75570            \\\hline
GPI-5710001000      &Alprazolam              &44214           \\\hline
GPI-5816007010      &Sertraline HCl          &39258            \\\hline
GPI-6420001000      &Acetaminophen           &20618            \\\hline
\end{tabular} \label{tab:DrugSelection1}
\end{table}

% \begin{table}[h]
% \caption{target Drugs in Group B}
% \centering  
% \begin{tabular}{|p{2.5cm}|p{3cm}|p{2cm}|}
% \hline
% \multicolumn{1}{|l|}{\textbf{Drug GPI code}} &
% \multicolumn{1}{l|}{\textbf{Description}}&
% \multicolumn{1}{l|}{\textbf{Patient Size}}
% \\ \hline
% GPI-7310001010      &Benzatropine   &4456\\    \hline
% GPI-7320307010      &Ropinirole     &7524\\ \hline
% GPI-7320306010      &Pramipexole    &4664\\\hline
% GPI-7315303000      &Entacapone     &445\\\hline
% GPI-7320001010      &Amantadine     &2597\\\hline
% GPI-7330003010      &Selegiline     &357\\\hline
% GPI-7320002010      &Bromocriptine  &272 \\\hline
% GPI-7320990210      &Levodopa       &6973\\\hline
% GPI-3040202000      &Cabergoline    &221\\\hline
% GPI-6205355010      &Memantine      &13171\\\hline
% \end{tabular} \label{tab:DrugSelection2}
% \end{table}

\mund{target ADEs}
ADEs are prevalent, and are not totally avoidable. The evaluation is performed on target ADEs that are severe. 
Table \ref{tab:ADESelection} shows the target ADE list used in evaluation, selected based on its severity according to existing studies  \cite{gottlieb2015ranking} and their occurrence in  our data set. Here the occurrence means the number of the patients experienced this ADE in our data set. Other ADEs can also be used in evaluation.

%ADEs are prevalent, and are not totally avoidable. Also, a large number of variables contribute to ADE occurrence in patients, it is impossible to precisely predict every ADE in every patient. Therefore, patient ADE risk prediction should focus on some specific ADEs in patients. In addition, since the aim is to prevent patient harm, it seems more rational to predict clinically relevant, potentially preventable adverse events. When a physician makes a prescription, it is important to know the risks of a patient experiencing severe ADE if the drug is taken.
%When a physician makes a prescription, it is important to know the risks of a patient experiencing severe ADE if the drug is taken. We collaborated with clinical expert in Inovalon to design a target ADE list based its severity and frequency. The ADE severity is followed by \cite{gottlieb2015ranking} and the frequency is calculated based on our data set. Table \ref{tab:ADESelection} shows the target ADE list used in evaluation. 

\begin{table}[ht]
\caption{Target Adverse Drug Events (ADEs)}
\centering  
\begin{tabular}{|p{3cm}|p{4cm}|}
\hline
\multicolumn{1}{|l|}{\textbf{ADE code (ICD 10)}} &
\multicolumn{1}{l|}{\textbf{Description}}\\ \hline
L29.9       & Pruritis \\    \hline
K27.9       & Stomach or intestinal ulcers    \\ \hline
L50.9       &Urticaria   \\\hline
T78.40      &Allergic Reaction   \\\hline
F329        &Depression \\\hline
R06.00      &Dyspnea  \\\hline
D649        &Anemia   \\\hline
D696        &Thrombocytopenia  \\\hline
M25.50      &Arthralgia   \\\hline
R00.2       &Palpitation  \\\hline
R20.2       &Paresthesia  \\\hline
F419        &Anxiety   \\\hline
M79.1       &Myalgia  \\\hline
I47.2       &Ventricular tachycardia  \\\hline
I63.0       &Anorexia\\\hline
\end{tabular} \label{tab:ADESelection}
\end{table}

\mund{Training and Testing Data}
For each target drug, we extract  all the patients whose claim history contains the GPI code of the drug. 
% There are  300K  unique patients take at least one drug in target drugs. 
% Note that a patient may take multiple target drugs.
% There are  about 0.8 million drug-patient instances in Group. 
For each patient, we extract the claim history before taking the target drug.
Then we identify the occurrence of a target ADEs within 3 months after the drug taking  using the method discussed in Section \ref{sec:prob} to generate the label for this instance.
We split all the patients in each drug into training/testing/validation dataset with ratio 0.7/0.2/0.1. The final result is the averaged result of these 10 drugs

\begin{table}[ht]
\small
\caption{Evaluation of Overall Effectiveness}
\centering  
\begin{tabular}{|p{2cm}|p{1cm}|p{1cm}|p{1cm}|p{1cm}|p{1cm}}
\hline
\multicolumn{1}{|l|}{\textbf{Systems}} & \multicolumn{1}{l|}{\textbf{Accuracy}} & \multicolumn{1}{l|}{\textbf{Precision}} & 
\multicolumn{1}{l|}{\textbf{Recall}} & 
\multicolumn{1}{l|}{\textbf{AUC}}\\ \hline
Random Forest   & 0.78      &0.65       &0.21       &0.75 \\    
XGBoost         & 0.80      &0.67       &0.25       &0.76 \\   
LSTM            & 0.84      &0.69       &0.34       &0.81 \\ 
CNN             &0.83       &0.60       &0.37       &0.80\\
\hline
\textbf{HTNNR} &\textbf{0.88}   &\textbf{0.84}  &\textbf{0.51}   &\textbf{0.89}\\
\hline
\end{tabular} \label{tab:OverallgroupA}
\end{table}

\subsection{Comparison Methods}

Since we are the only study that uses claims  history for personalized  ADE risk prediction,  there is no existing work to compare. We use several baseline approaches for comparison.  

\begin{itemize}
    \item\textbf{Long Short Term Memory (LSTM):} This is the method discussed in  Section  \ref{sec:method1}.
    
    \item\textbf{Convolutional Neural Network (CNN):} This replaces the LSTM model in the method discussed in  Section  \ref{sec:method1} with a CNN model. CNN has proven effectiveness in computer vision \cite{khan2018guide}, natural language processing \cite{yin2017comparative}  
    \item\textbf{Random Forest:} Random forest is a classification algorithm consisting of many decisions trees  \cite{liaw2002classification}. 
    \item\textbf{XGBoost:} XGBoost is an implementation of gradient boosted decision tree algorithm which has been widely used in many classification tasks like emotion analysis \cite{jabreel2018eitaka} and image classification \cite{ren2017novel}
\end{itemize}

Note that every method is trained on the patients for each target drug independently. For Random Forest and XGBoost, we use Term Frequency (TF)-Inverse Document Frequency (IDF) vectors extracted from claim code sequence as features. TF-IDF has been commonly used as features in text classification tasks \cite{zhang2011comparative}.

\subsection{Evaluation of Overall Effectiveness}
Table \ref{tab:OverallgroupA} shows the performance of different methods on target drugs. For each system, each number is the average performance  on ten drugs in each drug group.Several observations can be made.

The proposed HTNNR method consistently achieves the best performance among these methods on all metrics. One reason is that the hierarchical attention model to differentiate the  relationship of claim codes in an encounter, and the relationship of encounters in an claim history. It further uses different neural networks, Bi-LSTM, and TLSTM, respectively, to capture their different characteristics. 
On the other hand, comparison systems model the input as a sequence of claims code for each patient.
Furthermore, the attention layer in HTNNR gives higher weights on  important claim codes and important encounters. 

We also observe that the performance differences on precision and recall are much bigger than those on AUC and Accuracy. It is relatively easy for a model to perform well on AUC and Accuracy on  imbalanced data. AUC represents the model overall classification ability on various thresholds. It does not reflect well the effect of minority class. Even if a method mis-classifies most or all of the minority class, its AUC value can still be high. Similarly, for imbalanced data, if a model always predicts  the majority label, it will obtain a good accuracy. In our case, the target drug list has about 80\% negative labels. Thus most methods perform similarly on AUC and Accuracy. High AUC and Accuracy can be misleading in some imbalanced data. On the other hand, achieve high precision and recall are much more challenging. In the following, we focus the analysis on precision and recall.

\subsection{Evaluation on single drug}
\begin{figure}[ht]
  \includegraphics[width=\linewidth]{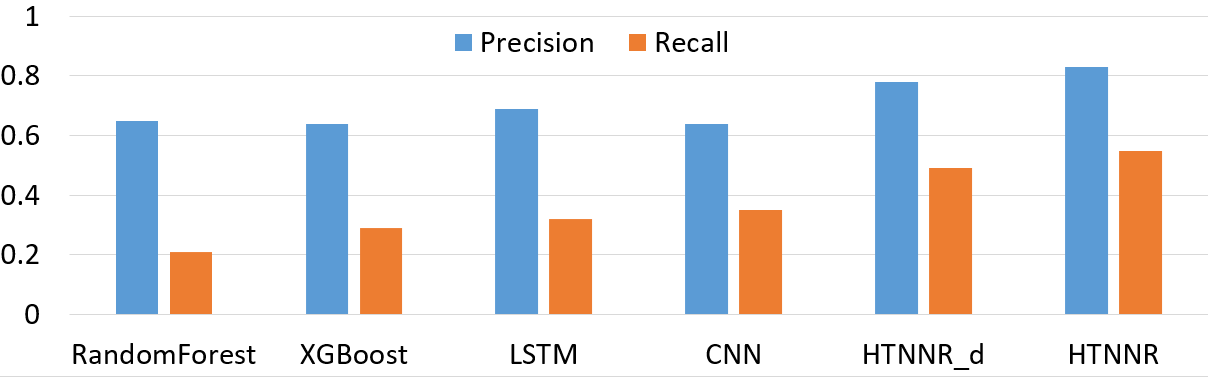}
  \caption{GPI-3320003010 (\# of Patients Taken: 75,561)}
  \label{fig:drugA}
\end{figure}

Table~\ref{tab:OverallgroupA} shows the average results on the 10 drugs. Now we zoom in to a single drug.
We randomly select a drug from target durg list,  GPI-3320003010, and evaluate the performance of comparison systems, and HTNNR on its ADE risk prediction, as shown in  Figure \ref{fig:drugA}.
Here we only show the precision and recall,  as the performance differences of Accuracy and AUC are similar as the  result represented in Table \ref{tab:OverallgroupA}. There are several things worth mention. 
First,  the  HTNNR model  performs better than the comparison systems, consistent with the evaluation shown in Table  \ref{tab:OverallgroupA}.

we also observe the improvement on recall is higher than that on precision. Hierarchical framework helps to find  more shared ADE characteristics among the drugs. At the same time, more noisy information is  introduced. Thus, the recall benefits more from  training data from multiple drugs than precision. 

\subsection{Evaluation of Patient Claim History Length}
\begin{figure}[ht]
  \includegraphics[width=\linewidth]{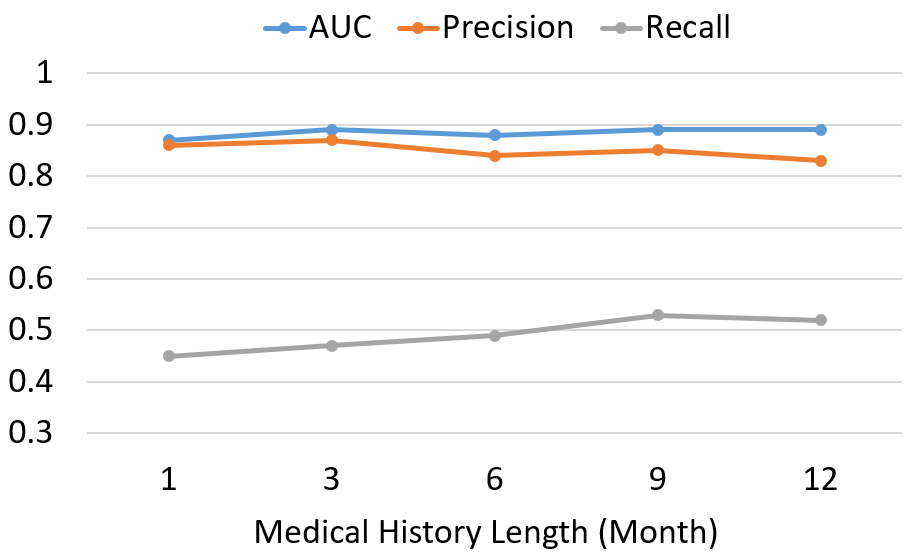}
  \caption{Evaluation  on Different Length of Claim History}
  \label{fig:historyLengthA}
\end{figure}

All the results shown so far takes as input 
each patient's entire  medical claim history before taking a target drug to train each  model. To evaluate how the patient claim history impact the personalized ADE prediction,  we evaluated the performance on different time length of medical history considered. 
Figure \ref{fig:historyLengthA} shows the  performance vary with varying length of each patient's medical history  used to train HTNNR.
The medical history always ending at the time when a target drug is recorded in the claim history, with duration count backward. The results show that using 3 month of claim history generates better performance than using 1 month of claim history, since the model can benefit from a larger dataset. After 3 months, 
the longer history considered, the better recall, and the worse precision. The reason is that longer  history data can help the model to find more characteristics of patients and potential drug interactions, but at the same time,  introduce more noisy information. In a real application, we can adjust the history length to be considered depends on which metrics is more important in the application.

To summarize,  HTNNR achieves the best effectiveness in all evaluation metrics among all methods tested. 

\section{Conclusions and Future Work}\label{sec:futurework}
In this paper,  we studied how to use patient claims history for personalized ADE risk prediction. We propose the HTNNR model  that captures the characteristics of claim codes and their relationship. It has a hierarchical framework. The first layer first generates embedding for claim codes, and then generate a vector for  each encounter using a Bi-LSTM model with an attention layer. The second layer takes the sequences of encounter vectors as input and uses a  time-aware neural network model to generate claim history representation that capture the temporal order of encounters. The empirical evaluation show that the proposed HTNNR model is  effective and efficient, especially for rare drugs.

Since claim history is updated on daily basis, as future work we will investigate how to incrementally train the model based on the new information available without re-training the model from scratch every time.

\bibliographystyle{IEEEtran}
\bibliography{reference}

\end{document}